\documentclass{article} % For LaTeX2e
\usepackage{main,times}

\usepackage{amsmath,amsfonts,bm}

\def\eqref#1{equation~\ref{#1}}
\def\1{\bm{1}}

\DeclareMathAlphabet{\mathsfit}{\encodingdefault}{\sfdefault}{m}{sl}
\SetMathAlphabet{\mathsfit}{bold}{\encodingdefault}{\sfdefault}{bx}{n}

\usepackage{hyperref}
\usepackage{url}
\usepackage{graphicx}
\usepackage{booktabs}
\usepackage{multirow}
\usepackage{tabularx}
\usepackage{makecell}
\usepackage{bbding}
\usepackage[printonlyused]{acronym}
\title{
A Survey on LLM Mid-Training
}

\author{
}

\author{
    Chengying Tu\textsuperscript{\rm 1,\rm 2}$^{\ast}$\quad
    Xuemiao Zhang\textsuperscript{\rm 1,\rm 2}\thanks{Equal contribution.}\quad
    Rongxiang Weng\textsuperscript{\rm 2}$^{\dagger}$\quad
    Rumei Li\textsuperscript{\rm 2} \\
    \bf Chen Zhang\textsuperscript{\rm 2}\quad
    Yang Bai\textsuperscript{\rm 2}\quad
    Hongfei Yan\textsuperscript{\rm 1}\thanks{Corresponding author.}\quad
    Jingang Wang\textsuperscript{\rm 2}\quad
    Xunliang Cai\textsuperscript{\rm 2}\\
    \textsuperscript{\rm 1} Peking University\quad
    \textsuperscript{\rm 2} Meituan \\
    \texttt{tuchengying@stu.pku.edu.cn}\quad
    \texttt{\{zhangxuemiao, yanhf\}@pku.edu.cn} \\
    \texttt{wengrongxiang@gmail.com} \quad
    \texttt{\{lirumei, zhangchen76\}@meituan.com} \\
    \texttt{\{baiyang28, wangjingang02, caixunliang\}@meituan.com}
}

\acrodef{LLM}[LLM]{Large language model}
\acrodef{LLMs}[LLMs]{large language models}
\acrodef{QA}[QA]{question-answering}
\acrodef{CoT}[CoT]{chain-of-thought}

\iclrfinalcopy % Uncomment for camera-ready version, but NOT for submission.
\begin{document}

\maketitle

\begin{abstract}

Recent advances in foundation models have highlighted the significant benefits of multi-stage training, with a particular emphasis on the emergence of mid-training as a vital stage that bridges pre-training and post-training. Mid-training is distinguished by its use of intermediate data and computational resources, systematically enhancing specified capabilities such as mathematics, coding, reasoning, and long-context extension, while maintaining foundational competencies. This survey provides a formal definition of mid-training for large language models (LLMs) and investigates optimization frameworks that encompass data curation, training strategies, and model architecture optimization. We analyze mainstream model implementations in the context of objective-driven interventions, illustrating how mid-training serves as a distinct and critical stage in the progressive development of LLM capabilities. By clarifying the unique contributions of mid-training, this survey offers a comprehensive taxonomy and actionable insights, supporting future research and innovation in the advancement of LLMs.

\end{abstract}

\section{Introduction\label{sec:introduction}}
The paradigm shift in foundation model development has transitioned from monolithic pre-training approaches to sophisticated multi-stage optimization frameworks~\citep{ibrahim2024simple, blakeney2024does, feng2024maximizedataspotentialenhancing, zhang-etal-2025-frame, zhang-etal-2025-preference}. While general pre-training establishes fundamental competencies through exposure to diverse large-scale corpora, contemporary research demonstrates that subsequent optimization phases systematically amplify specialized capabilities like mathematics, reasoning, coding, agent, and long-context extension~\citep{grattafiori2024llama3herdmodels, parmar2024nemotron415btechnicalreport, olmo20252olmo2furious}. This evolution reflects a growing consensus that general pre-training may not effectively or sufficiently cultivate the capabilities required in specialized domains, particularly those that demand sustained access to high-quality data sources. The demonstrated potential of intermediate optimization phases has catalyzed their formalization as a distinct developmental stage, which is now gradually being recognized as the \textit{mid-training} stage.

\textit{Mid-training is positioned as the critical bridge between general pre-training and post-training stages}, characterized by intermediate computational demands and targeted large-scale data utilization. The mid-training stage has proven its capacity for bidirectional capability balance: forward-propagating specialized capabilities potential through curriculum-guided exposure to domain-specific data, while simultaneously backward-preserving general competencies via a reserved general data ratio. \textit{While pre-training focuses on establishing foundational capabilities, mid-training aims to preserve these foundations while amplifying targeted competencies.} Empirical evidence indicates that mid-training delivers steeper performance gains with less data and computation than pre-training~\citep{wang2025octothinker,meituanlongcatteam2025longcatflashthinkingtechnicalreport}.

Mid-training encompasses a range of operations designed to enhance model capabilities, although it currently lacks systematic organization. For example, during the annealing phase, characterized by fast learning rate decay, the strategic introduction of high-quality data enables models to converge toward superior local optima~\citep{hu2024minicpmunveilingpotentialsmall}. Notably, for challenging tasks like mathematics and coding, exclusive reliance on post-training stages risks data distribution shifts that undermine general capabilities, whereas integration during the annealing phase proves more effective~\citep{gunter2024appleintelligencefoundationlanguage,akter2025frontloadingreasoningsynergypretraining}. Similarly, effective context window expansion often demands modifying training objectives in later stages while concurrently increasing the proportions of long-context data~\citep{grattafiori2024llama3herdmodels}. Furthermore, achieving multilingual proficiency requires elevating the ratios of low-resource language data~\citep{abdin2024phi3technicalreporthighly, wake2025yilightningtechnicalreport}.

This survey systematically formalizes the concept of mid-training in LLM development and establishes a structured taxonomy of optimization frameworks specifically tailored for this critical stage. Our analysis reveals that mid-training optimization demonstrates unique characteristics that balance computational efficiency, model stability, and capability enhancement. This survey presents three key contributions: (1) A rigorous definition of mid-training contextualized within its historical development; (2) A comprehensive optimization framework spanning data curation, training strategies, and model architecture; (3) A review of objective-oriented implementations in mainstream models.

The subsequent sections unfold through three integrated perspectives. First, we elaborate on the conceptual definition and trace the evolution of mid-training (Section~\ref{sec:definition}). Second, we develop a unified optimization framework comprising five interconnected components: data curation (Section~\ref{sec:data}), training strategies (Section~\ref{sec:training}), model architecture optimization (Section~\ref{sec:model}), decay scaling laws (Section~\ref{sec:scalinglaw}), and evaluation (Section~\ref{sec:evaluation}). Third, we present the specific objectives of mid-training and detail the concrete optimization implementations adopted by mainstream models to achieve these objectives (Section~\ref{sec:objectives}).

\section{Overview\label{sec:definition}}

\textbf{The concept of mid-training remains empirically valuable yet theoretically nebulous, as current literature employs it heterogeneously to describe targeted interventions during foundation model development.} While phi-3.5~\citep{abdin2024phi3technicalreporthighly} operationalizes mid-training as a phase for multilingual and long-context data integration, Yi-Lightning~\citep{wake2025yilightningtechnicalreport} conceptualizes it as a transitional period governed by controlled data distribution shifts. OLMo 2~\citep{olmo20252olmo2furious} further diverges by associating mid-training with advanced curriculum learning techniques, particularly emphasizing learning rate annealing and data mixing strategies. This terminological inconsistency reveals a critical gap in the field—the absence of a rigorous taxonomy to unify mid-training’s objectives, methodologies, and boundaries within the model development lifecycle.

\begin{figure}[t]
\centering
\includegraphics[width=\columnwidth]{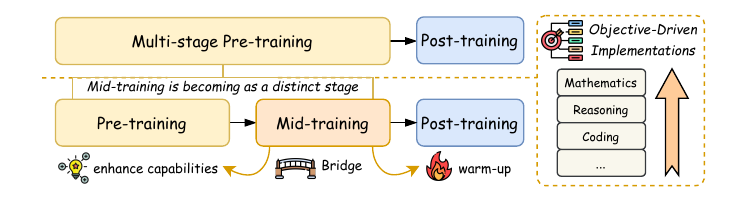}
\label{fig:position}
\caption{Mid-training has emerged as a distinct stage, separate from conventional multi-stage pre-training paradigms. Serving as a bridge between pre-training and post-training stages, mid-training enhances the specialized capabilities of pre-trained models while providing a critical transition and warm-up for subsequent post-training stages. Mid-training encompasses multiple objective-driven implementations, spanning domains such as mathematics, reasoning, coding, and more.}
\end{figure}

\textbf{Mid-training occupies a distinct functional position in the training pipeline, situated between general pre-training and specialized post-training stages.} As illustrated in Figure~\ref{fig:position}, traditional multi-stage training methodologies progressively enhance model performance and generalization by adjusting learning rates and data distributions across different phases~\citep{zhang-etal-2025-frame}. The curriculum learning paradigm naturally aligns with the multi-stage pre-training data curation framework, emphasizing the strategic ordering of training data based on criteria such as data quality, difficulty level, and domain mastery~\citep{nair-etal-2024-curriculum, zhang-etal-2025-preference}. 
Due to its broad research value and potential, mid-training is increasingly being recognized as a distinct stage, separate from multi-stage training, serving not only as a bridge in terms of data distribution~\citep{liu2025midtrainingbridgespretrainingposttraining} but also enabling targeted adjustments to training strategies and model architectures aimed at developing specific capabilities.
Within this emerging classification framework, contemporary LLM development designate \textit{pre-training as the foundational stage}, where models acquire general capabilities by predicting the next token over large-scale corpora. Emerging as an evolutionary bridge, \textit{mid-training} preserves these foundational capabilities with the same next-token prediction objective, while systematically enhancing specialized skills through curated data blends—often incorporating high-quality domain-specific content and instruction data. Unlike pre-training, mid-training demonstrates steeper performance gains with higher data efficiency, frequently employing multiple sub-phases for different capability upgrades. In contrast, post-training adopts specialized objectives for SFT/RL and task-specific alignment datasets.

As a critical intermediary, mid-training performs bidirectional mediation—\textbf{it enhances specialized capabilities of the pre-trained model while providing essential transition and warm-up for subsequent post-training stages.} It anchors three capability domains: core cognitive skills (e.g., mathematics/STEM, reasoning), task execution (e.g., instruction following, coding, agent behaviors), and extensibility (e.g., long-context processing, multilingualism). The optimization framework encompasses strategic data curation, training strategies (e.g., learning rate scheduling), and model architecture optimization (e.g., modifications for long-context extension). These components are intricately integrated to serve specific objectives.  For example, to achieve long-context extension, it necessitates not only the incorporation of high-quality long-context data but also structural adjustments to the model itself to ensure effective handling of extended contextual information. 

Mid-training might be confused with continued pre-training, which extends pre-training with domain-specific data without regard to original optimizer states or distribution preservation~\citep{gururangan-etal-2020-dont}. The former represents a deliberate developmental stage of the foundation model with transitional intent, \textbf{consciously blending pre-training data proportions and inheriting its learning rate dynamics to prevent catastrophic forgetting}. The latter may lead to the loss of the model's general capabilities.

% \section{Core Operational Toolkit\label{sec:toolkit}}

% This section delineates the core operational toolkit for mid-training, encompassing toolkit for data curation, specialized training strategies, decay scaling laws, and capability-centric evaluation frameworks, some of which exhibit cross-stage applicability.

\section{Data Curation\label{sec:data}}

\begin{figure}[t]
\centering
\includegraphics[width=0.7\columnwidth]{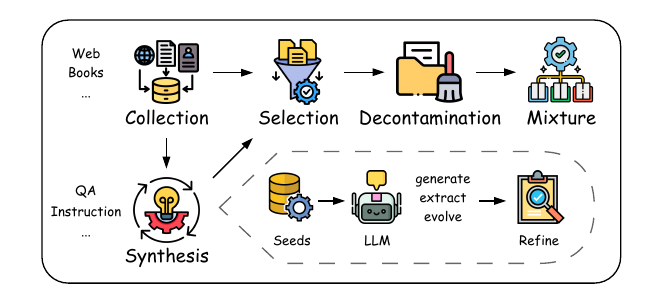}
\caption{Pipeline of mid-training data curation.}
\label{fig:data_curation}
\end{figure}

Mid-training data typically comprises a hybrid composition of general high-quality corpora and specialized formats such as QA pairs, instruction data, and domain-specific data, including mathematics and coding. This section systematically and comprehensively examines the end-to-end workflow of data curation for mid-training, detailing critical tools and methodologies spanning data collection, synthesis, selection, decontamination, and mixture, as shown in Figure~\ref{fig:data_curation}. The sequence of each component may be adjusted according to the specific construction requirements. For example, data decontamination can be performed hierarchically in multiple steps. Each component is addressed to elucidate its synergistic role in constructing robust mid-training datasets.

\subsection{Data Collection}

The sources of data collection for mid-training are largely analogous to those employed during pre-training, encompassing web-crawled corpora, digitized books, and human-annotated materials. The widely used web crawling tools include Trafilatura~\citep{barbaresi-2021-trafilatura}, and widely adopted datasets include CommonCrawl, Wikipedia, C4, Pile~\citep{gao2020pile800gbdatasetdiverse}, Dolma~\citep{soldaini-etal-2024-dolma}, RedPajama~\citep{weber2024redpajama}, and Matrix~\citep{zhang2024mapneohighlycapabletransparent}. In addition to these general sources, mid-training data also collects specialized data formats such as QA pairs, with datasets including Stack Exchange QA, RedStone~\citep{chang2024redstonecuratinggeneralcode}, MegaMath~\citep{zhou2025megamathpushinglimitsopen}, and so on. All collected data is subjected to preliminary cleaning and quality filtering to establish data integrity and reliability. The data collection process serves two primary objectives: firstly, to elevate the overall quality of the dataset while maintaining continuity with the pre-training data distribution, thereby alleviating the risks associated with distributional shift;   and secondly, to facilitate the synthesis of task-specific datasets required for mid-training, such as the generation of QA pairs through techniques like rephrasing.   This systematic approach to data collection and preparation ensures that mid-training is supported by both high-quality and appropriately distributed data, laying a robust foundation for subsequent model specialized capabilities enhancement.

\subsection{Data Synthesis}

Data synthesis serves as a powerful tool for reconstructing and transforming existing datasets in style or format, and generating scarce types of natural data such as agentic data, thereby enhancing information density and overall data quality. It effectively addresses critical challenges in mid-training, such as data scarcity, insufficient diversity, and quality limitations, by enabling robust data augmentation and enrichment. The strategic importance of synthetic data is increasingly recognized. For instance, phi-4~\citep{abdin2024phi4technicalreport} highlights the advantages of synthetic data, including structured and gradual learning as well as alignment with inference contexts. Furthermore, the domain-specific, high-quality data utilized by OLMo 2~\citep{olmo20252olmo2furious} during mid-training is primarily synthetic.

Given the large scale of data involved in mid-training, most of the currently popular synthesis methods are LLM-driven and can be broadly categorized into three types: distillation, extraction, and evolution.

\textbf{Distillation} leverages powerful LLMs to directly generate data through carefully engineered prompts or to distill smaller synthesis models. This approach is heavily reliant on the knowledge and capabilities of LLMs and is often followed by rigorous quality filtering or refinement.

(1) Style rephrasing techniques transform low-quality, noisy corpora into higher-density expressions. Rephrased web data demonstrates superior utility by aligning with downstream stylistic diversity, such as QA, instruction, and dialogue, and by enhancing overall data quality~\citep{maini2024rephrasing, akter2025mind}.

(2) Diffusion synthesis for specific formats typically involves designing specified prompts and generating massive QA pairs or instruction data based on corpora~\citep{cheng2024instruction, chen2024towards}. Alternatively, it may leverage intermediaries such as personas to enhance diversity and scale~\citep{ge2025scalingsyntheticdatacreation}. 
Building upon these insights, BoostQA~\citep{zhang2025largescalediversesynthesismidtraining} identifies deficiencies in model performance related to STEM disciplines and high-difficulty tasks through probe experiments and proposes a novel large-scale diverse synthesis pipeline focused on STEM and high-difficulty data. 
Furthermore, LinkQA~\citep{zhang2025linkqasynthesizingdiverseqa} proposes a novel strategy that leverages path sampling on the knowledge point graph to construct multi-seed QA synthesis. 
Similarly, code synthesis utilizes designed prompts to generate code instructions based on open-source code snippets~\citep{luo2024wizardcoder, wei2024magicoder}, or abstracts code into concepts for instruction construction~\citep{wei2024selfcodealign}.

(3) LLM-driven low-resource language translation and multimodal synthesis—such as VLM-driven PDF parsing for data generation~\citep{yang2025qwen3technicalreport}—extend the utility of data synthesis. Crucially, the core principles underlying these methods are transferable to diverse data formats, providing systematic inspiration for efficient, high-quality synthetic data pipelines during mid-training.

\textbf{Extraction} leverages LLMs to directly extract natural QA pairs or other data formats from collected corpora, followed by subsequent refinement. For example, WebInstruct~\citep{yue2024mammoth} implements a pipeline of recall, extract, and refine to harvest high-quality QA pairs from web content, while MegaMath-Synthetic~\citep{zhou2025megamathpushinglimitsopen} combines QA extraction with solution refinement to improve diversity and quality of mathematical reasoning data.

\textbf{Evolution} generates enhanced problems and solutions through designed cyclic iterations, especially for mathematics. For instance, MathGenie~\citep{lu-etal-2024-mathgenie} augments solutions from seed data and back-translates them into novel mathematical problems, and subsequently generates verified answers, creating diverse and reliable synthetic data.

These methodologies collectively demonstrate the versatility and transformative potential of synthetic data in overcoming natural data limitations.

\subsection{Data Selection}

Data selection during mid-training involves filtering high-quality or domain-specific samples from raw data to derive target datasets. This process encompasses both general and synthetic data sources and operates at a finer granularity than the pre-training stage. Data selection is primarily achieved through two methodological approaches: targeted sampling and rater-based filtering.

Targeted sampling adjusts data distribution by downsampling less relevant domains or upsampling domain-specific content to enhance representation. Rater-based filtering employs specialized scoring models to evaluate data quality, including the FastText classifier~\citep{joulin-etal-2017-bag}, a binary classifier used to distinguish data quality; FineWeb-Edu~\citep{penedo2024the}, a classifier that assigns scores based on adherence to academic topics and polished content; and QuRater~\citep{wettig2024qurating}, a rater model that assigns ratings corresponding to pairwise judgments.

The current bottleneck in mid-training data selection lies in filtering pre-training data into high-quality general datasets, predominantly through synergistic combinations of multiple raters~\citep{wettig2024qurating, duan2025enhancing, xu2025fireflexibleintegrationdata} to cherry-pick optimal samples. Conversely, synthetic or structurally specialized data benefits from more targeted filtering strategies~\citep{zhou2024jiuzhang}, which enable the efficient extraction of domain-aligned, premium content.

\subsection{Data Decontamination}

Data decontamination has emerged as a critical preprocessing step for foundation models, aiming to eliminate nonsense, sensitive, or benchmark-related content from training corpora. This process mitigates data leakage risks and ensures fair evaluation of model capabilities. The primary challenge lies in balancing the removal of contaminated data while preserving model performance, as existing methods inherently suffer from false positives—over-aggressive filtering that degrades generalizability, and false negatives—incomplete decontamination that inflates benchmark scores~\citep{singh2024evaluationdatacontaminationllms}. 

Current approaches remain largely empirical, with no theoretically optimal solution to reconcile these trade-offs. N-gram matching dominates practical implementations due to its simplicity and scalability. This method can be used to remove potentially contaminated samples from benchmarks~\citep{brown2020languagemodelsfewshotlearners, grattafiori2024llama3herdmodels} and also to remove possibly contaminated data from training corpora~\citep{abdin2024phi4technicalreport}, with the choice of $n$ significantly impacting effectiveness. Despite its prevalence, n-gram matching fails to capture semantic equivalences through lexical variations (e.g., synonym substitution) and relies heavily on heuristic parameter tuning. Embedding-based methods, such as cosine similarity comparisons using pretrained language models, offer enhanced semantic sensitivity but incur prohibitive computational costs for large-scale datasets. Hybrid methods, such as n-gram filters combined with additional longest common subsequence criteria, demonstrate promising efficacy gains but lack universal applicability~\citep{yang2024qwen25mathtechnicalreportmathematical}.

\subsection{Data Mixture}

Mid-training strategically integrates diverse data forms to enhance specified capabilities. The data composition typically blends high-quality general corpora with specialized data formats, such as QA and instruction data, with ratios tailored to specific objectives~\citep{hu2024minicpmunveilingpotentialsmall, allenzhu2024physicslanguagemodels31, wake2025yilightningtechnicalreport}. High-quality general corpora, usually filtered from pre-training data, maintain foundational linguistic robustness and mitigate distribution shift risks~\citep{ibrahim2024simple, blakeney2024does}. Prior experiments demonstrate that relying solely on synthetic data exacerbates hallucinations and degrades performance on knowledge-intensive tasks, underscoring the necessity of general data for stability~\citep{abdin2024phi4technicalreport}.

Specialized data formats are introduced or augmented during mid-training as a warm-up to align with downstream tasks, each serving distinct purposes.

\textbf{QA data} is characterized by knowledge-intensive or reasoning-driven question-answer interactions. These pairs excel at structuring explicit knowledge representations and enhancing few-shot adaptability, with studies demonstrating their superiority in evaluation scenarios through format diversity~\citep{allenzhu2024physicslanguagemodels31}. While early research often treats QA as a monolithic category, its granular subtypes can target distinct capabilities: answer-only or short-CoT QA prioritizes factual knowledge mastery, whereas long-CoT or reflective QA systematically cultivates iterative reasoning capabilities. The roles of different QA formats are still being explored.

\textbf{Instruction data} shares structural similarities with post-training instruction-tuning datasets. This data type often overlaps with QA formats in knowledge-intensive domains, where instructions may implicitly or explicitly query factual knowledge. Empirical studies demonstrate that the incorporation of a small amount of instruction data (e.g., 1\% of training samples) significantly reduces the amount of instruction tuning required during the SFT stage and enhances RL effectiveness~\citep{tencenthunyuanteam2025hunyuanturbosadvancinglargelanguage, wang2025octothinker}. Instruction data also includes content for specific domains, such as code instructions and agent workflow instructions.

\textbf{Long-context data} is characterized by extensive token sequences typically sourced from books, wikis, and papers with high quality. It is generally upsampled or augmented through synthesis to enhance the long-context processing capabilities of models. Furthermore, long-CoT QA data is also becoming an important component of long-context data~\citep{hu2024minicpmunveilingpotentialsmall}.

These diverse data formats are strategically employed to support specific model capability objectives, as discussed in Section~\ref{sec:objectives}. For instance, QA and instruction data are commonly used to facilitate knowledge mastery, enhance reasoning abilities, and enable specialized tasks such as code-related and agentic question answering. The final mid-training dataset is constructed through the careful integration of these diverse data formats, each targeted at specific training objectives.

Although the determination of optimal data mixing ratios in mid-training inherently depends on the pre-trained model's capabilities and target objectives, the methodologies for identifying these ratios exhibit broad applicability across different scenarios. Empirically driven allocations remain prevalent; for example, Llama-3 assigns a 30\% weight to novel datasets while reserving 70\% for its default pre-training data blend~\citep{grattafiori2024llama3herdmodels}. Beyond heuristic approaches, recent techniques systematically correlate mixing proportions with downstream performance. OLMo 2~\citep{olmo20252olmo2furious} employs microanneals, which abbreviate annealing runs on small subsets to extrapolate the efficacy of large-scale mixtures from small-scale trials. Concurrently, RegMix~\citep{liu2025regmix} formalizes ratio optimization as a regression task, training many small models across diverse mixing configurations to predict unseen mixture performance via regression modeling. These methods hinge on the stability hypothesis, which asserts that ratio performance rankings persist across model scales and token volumes, thereby enabling computationally efficient generalization from proxy experiments to production-level training pipelines.

\section{Training Strategies\label{sec:training}}

The training strategies for mid-training primarily focus on learning rate (LR) schedules, while other hyperparameter configurations also significantly impact training outcomes.

\subsection{Learning Rate Scheduling\label{sec:lr}}

The LR scheduling is a critical component in the mid-training stage, influencing the stability, efficiency, and overall performance of model training. As illustrated in Table~\ref{tab:lr}, LR scheduling typically consists of a warm-up phase followed by a decay phase. 
During the warm-up phase, linear warm-up is commonly employed, gradually increasing the LR from zero or a specified value to its peak value over a designated proportion of training steps. Empirical research has demonstrated that this warm-up phase can improve training stability~\citep{gilmer2022a}.
In the decay phase, various strategies such as linear, cosine, and exponential decay can be applied. The decay phase is often divided into a gradual decay period and a fast decay period, with high-quality data commonly introduced during the fast decay stage~\citep{grattafiori2024llama3herdmodels}.
In addition, multi-stage schedulers such as the Warmup-Stable-Decay (WSD) scheduler~\citep{hu2024minicpmunveilingpotentialsmall} synchronize LR adjustments with distinct training stages, enabling the LR to better accommodate evolving data distributions and shifting optimization objectives throughout the training process. WSD introduces a stable training phase with a constant high LR, which helps the model to explore the parameter space and potentially find a better global optimum. This approach enhances both optimization efficiency and generalization performance, and is therefore widely adopted in current practice.

\begin{table}[t]
    \caption{Different LR schedulers.}
    \label{tab:lr}
    \begin{center}
    \small
    \begin{tabular}{l|l|l}
        \toprule
        \multicolumn{2}{c}{\bf LR Scheduler} & \bf Function Form \\
        \midrule
        \bf Warmup & Linear & $\eta=\frac{s}{W}{\eta}_{\max}, \quad 0\le s\le W$ \\
        \midrule
        \bf Constant & - & $\eta=\eta, \quad 0\le s\le T$ \\
        \midrule
        \multirow{3}{*}{\bf Decay} & Linear & $\eta = {\eta}_{\max} - ({\eta}_{\max} - {\eta}_{\min})\cdot\frac{s}{S},\quad 0\le s\le S$ \\
        & Cosine & $\eta = {\eta}_{\min} + \frac{1}{2}({\eta}_{\max}-{\eta}_{\min})(1+\cos(\pi\frac{s}{S})), \quad 0\le s\le S$ \\
        & Expotential & $\eta={\eta}_{\max}\cdot e^{-ks}$  \\
        \midrule
        \bf Multi-stage & WSD & $WSD(T;s)=\begin{cases}
            \frac{s}{W}\eta, \quad s<W \\
            \eta, \quad W<s<T \\
            f(s-T)\eta, \quad T<s<S
        \end{cases}$ \\
        \bottomrule
    \end{tabular}
    \end{center}
\end{table}

\subsection{Other Training Settings\label{sec:othersettings}}

The key hyperparameters typically configured alongside LR scheduling include batch size. Increasing the batch size reduces the variance in stochastic gradient estimates, which, in turn, allows for the adoption of higher LR during training. Empirical studies~\citep{goyal2018accuratelargeminibatchsgd} have shown that, particularly in large-scale mini-batch scenarios, employing a conservative initial learning rate combined with a well-designed warm-up phase can alleviate optimization challenges. During the annealing phase, the batch size is usually influenced by the data scale and may be dynamically adjusted~\citep{grattafiori2024llama3herdmodels}.

\section{Model Architecture Optimization\label{sec:model}}

\subsection{Long Context Extension\label{sec:long_context}}

Rotary Position Embedding (RoPE)~\citep{roformer} has become the standard position embedding in LLMs. However, LLMs pre-trained with fixed context length $L$ might face significant performance degradation when processing longer sequences, necessitating the use of long-context extension methods built upon RoPE. Herein we include several commonly used extension variants.

Before diving into these extension methods, we first briefly describe RoPE itself. RoPE encodes the position information of tokens with a rotation tensor that naturally incorporates explicit relative position dependency. To illustrate, given a hidden vector $h=[h_0,h_1,...,h_{d-1}]$, where $d$ is the hidden dimension, and a position index $m$, RoPE operates as follows:
\begin{equation}
\resizebox{0.89\linewidth}{!}{$
	f(h,m) = 
	\begin{pmatrix}
		h_0\\
		h_1\\
		h_2\\
		h_3\\
		\vdots\\
		h_{d-2}\\
		h_{d-1}
	\end{pmatrix}
	\otimes
	\begin{pmatrix}
		\cos{m\theta_0} \\
		\cos{m\theta_0} \\
		\cos{m\theta_1} \\
		\cos{m\theta_1} \\
		\vdots \\
		\cos{m\theta_{d/2-1}} \\
		\cos{m\theta_{d/2-1}} 
	\end{pmatrix}
	+
	\begin{pmatrix}
		-h_1\\
		h_0\\
		-h_3\\
		h_2\\
		\vdots\\
		-h_{d-1}\\
		h_{d-2}
	\end{pmatrix}
	\otimes
	\begin{pmatrix}
		\sin{m\theta_0}\\
		\sin{m\theta_0}\\
		\sin{m\theta_1}\\
		\sin{m\theta_1}\\
		\vdots\\
		\sin{m\theta_{d/2-1}}\\
		\sin{m\theta_{d/2-1}}
	\end{pmatrix}$}
\label{eq_rope}
\end{equation} 
where $\theta_j=b^{-2j/d},j\in\{0,1,...,d/2-1\}$. 

\paragraph{Position Interpolation (PI)}

As described in~\citet{DBLP:journals/corr/abs-2306-15595}, PI involves proportionally down-scaling the position index $m$ to $m/\alpha$ in Equation \ref{eq_rope}.

\paragraph{NTK-aware Interpolation (NTK)}

NTK~\citep{rozière2023code} assumes that interpolating all dimensions equally, as done by PI, may result in the loss of high-frequency information. Therefore, NTK introduces a nonlinear interpolation strategy by adjusting the base frequency $b$.

\paragraph{Yet another RoPE extensioN (YaRN)}

Unlike PI and NTK, which treat each dimension of RoPE uniformly, YaRN~\citep{peng2023yarn} employs a ramp function to combine PI and NTK at varying proportions across different dimensions. Additionally, it introduces a temperature factor $t = \sqrt{1 + \ln(s)/d}$ to mitigate the distribution shift of the attention caused by long inputs.

RoPE extensions has evolved from simple interpolation to sophisticated frequency-aware methods. YaRN currently represents the optimal balance of performance and efficiency for production systems, while NTK with the dynamic adjustment offers the fastest path to zero-shot extension. The field continues to advance with hybrid approaches and learned scaling strategies.

\section{Decay Scaling Laws\label{sec:scalinglaw}}

Distinct from pre-training scaling laws, decay scaling laws account for the unique starting points of the decay phase, offering a tailored approach to predicting key variables that impact training efficacy. These laws primarily focus on the following predictive variables: model size, where the checkpoint remains fixed while the data ratio and training token count are specified; data ratio, which encompasses both general and specialized proportions, given the model size, checkpoint, and training token count; and training token count, provided the model size, checkpoint, and data ratio. The formulation of decay scaling laws addresses the complexities inherent in the decay phase, providing a structured framework for understanding the interplay between these factors and their influence on model performance. Further research is required to refine these predictions and fully elucidate the implications of decay scaling laws for optimizing model training processes.

\section{Evaluation\label{sec:evaluation}}

The evaluation of models during the mid-training stage remains an integral component of foundational model assessment, adhering to established standardized benchmarks as outlined in Table~\ref{tab:evaluation}. The evaluation framework encompasses multiple domains, including general, mathematical, coding, agent, and long-context domains, and is strategically aligned with the mid-training objectives detailed in Section~\ref{sec:objectives}.

\begin{table}[t]
    \caption{Mid-training benchmarks.}
    \label{tab:evaluation}
    \begin{center}
    \renewcommand{\arraystretch}{1.4}
    \small
    \begin{tabularx}{\textwidth}{l|l|X}
        \toprule
        \textbf{Domain} & \textbf{Abilities} & \textbf{Evaluation} \\
        \midrule
        General & Knowledge & MMLU~\citep{hendrycks2021measuringMMLU}, MMLU-Pro~\citep{wang2024mmlupro}, CMMLU~\citep{li-etal-2024-cmmlu}, C-Eval~\citep{huang2023ceval} \\
        \cline{2-3}
        & Reasoning & WinoGrande~\citep{sakaguchi2021winogrande}, HellaSwag~\citep{zellers-etal-2019-hellaswag}, ARC-C~\citep{clark2018think}, BBH~\citep{suzgun-etal-2023-challenging}, DROP~\citep{dua-etal-2019-drop}, PIQA~\citep{bisk2020piqa} \\
        \cline{1-3}
        Mathematical & Reasoning & GSM8K~\citep{cobbe2021training}, MATH~\citep{hendrycks2021measuringmath} \\
        \cline{1-3}
        Coding & Code Generation & HumanEval~\citep{chen2021evaluatinglargelanguagemodels}, MBPP~\citep{austin2021programsynthesislargelanguage}, MultiPL-E~\citep{cassano2022multiplescalableextensibleapproach}, DS-1000~\citep{lai2023ds}, LiveCodeBench~\citep{jain2024livecodebenchholisticcontaminationfree}, BigCodeBench~\citep{zhuo2025bigcodebenchbenchmarkingcodegeneration} \\
        \cline{2-3}
        & Code Completion & RepoBench~\citep{liu2023repobenchbenchmarkingrepositorylevelcode}, CrossCodeEval~\citep{ding2023crosscodeeval} \\
        \cline{2-3}
        & Code Fixing & SWE-bench~\citep{jimenez2024swebench}, OctoPack~\citep{muennighoff2023octopack},  Defects4J~\citep{Defects4J} \\
        \cline{2-3}
        & Code Reasoning & CruxEval~\citep{gu2024cruxevalbenchmarkcodereasoning} \\
        \cline{1-3}
        Agent & Tool Use & $\tau^2$-Bench~\citep{barres2025tau2benchevaluatingconversationalagents}, BFCL V3~\citep{patil2025the}, ACEBench~\citep{chen2025acebenchwinsmatchpoint} \\
        \cline{1-3}
        Long-context & Reasoning & LongEval~\citep{wu2025longevalcomprehensiveanalysislongtext}, MRCR~\citep{vodrahalli2024michelangelolongcontextevaluations}, LongBench v2~\citep{bai2025longbenchv2deeperunderstanding}, Ruler~\citep{hsieh2024ruler}, HELMET~\citep{yen2025helmetevaluatelongcontextlanguage} \\
        % \cline{1-3}
        % & Multilingual & - & \\
        \bottomrule
    \end{tabularx}
    \end{center}
\end{table}

\section{Objective-Driven Implementations\label{sec:objectives}}

Objective-driven implementations constitute the methodological cornerstone of mid-training, where capability enhancement objectives can be systematically categorized into the following interconnected domains: general capabilities, core cognitive abilities, task execution capabilities, and extension capabilities.
% , as well as efficiency optimization. 
In this section, we deconstruct the realization of these objectives through targeted intervention strategies. We clarify each objective in terms of its expected outcomes and analyze how mainstream models achieve these objectives by leveraging tailored approaches in data curation, training strategies, and model architecture optimization. Table~\ref{tab:targets} illustrates the mid-training objectives mentioned in the mainstream models. Furthermore, we discuss key insights and feasible directions for improvement, offering a foundation for further research and practical advancements in mid-training methodologies.

\subsection{General Capabilities}

During mid-training, maintaining and enhancing models' general understanding and generative capabilities emerges as a crucial optimization direction. This process necessitates carefully designed interventions to mitigate catastrophic forgetting, while simultaneously refining the model’s specialized competencies. As illustrated in Table~\ref{tab:targets}, all mainstream models dive deep into maintaining general performance while improving other objectives. Empirical analyses substantiate that retaining rigorously curated subsets of pre-training data is paramount, optimized composition and proportional balancing of which demonstrably elevate downstream task performance and generalization robustness~\citep{tencenthunyuanteam2025hunyuanturbosadvancinglargelanguage}. Methodologically, this manifests either through calibrated downsampling of general corpora to amplify specialized data proportions~\citep{gunter2024appleintelligencefoundationlanguage, huo2025dotsllm1technicalreport} or via further quality filtration of pre-training sources using advanced classifiers~\citep{parmar2024nemotron415btechnicalreport, olmo20252olmo2furious, tencenthunyuanteam2025hunyuanturbosadvancinglargelanguage}. 
For instance, BoostQA~\citep{zhang2025largescalediversesynthesismidtraining} and LinkQA~\citep{zhang2025linkqasynthesizingdiverseqa} facilitate rigorous quality stratification of pre-training datasets, using the QuRater model~\citep{wettig2024qurating} to quantifies knowledge density and the FineWeb-Edu educational classifier~\citep{penedo2024the} to assess educational utility. Texts rated highly in both knowledge density and educational utility are retained.
Collectively, these techniques ensure data continuity in the training process while leaving training space for specific data samples.

\begin{table}[t]
    \caption{Learning rate schedulers of mainstream models. The function form is shown in Table~\ref{tab:lr}. All linearly warm up from 0 unless specified, with the parameter representing the peak LR. For decay functions, the first parameter represents the end LR, and the second represents the token count. The subscript indicates the long-context extension.}
    \label{tab:lrModels}
    \begin{center}
    \small
    \begin{tabularx}{0.98\textwidth}{l|X}
        \toprule
        \bf Model & \bf Learning Rate Dynamics \\
        \midrule
        \makecell[lt]{dots.llm1 (2025/05) \\ \citep{huo2025dotsllm1technicalreport}} & Warmup(3e-4) $\to$ Constant(3e-4, 10T) $\to$ Decay(3e-5, 1T) $\to$ Decay(1e-5, 200B) $\to$ Constant(1e-5, 128B)$_{\text{8K} \to \text{32K}}$ \\
        \midrule
        \makecell[lt]{MiMo-7B (2025/05) \\ \citep{xiaomi2025mimounlockingreasoningpotential}} & Warmup(1.07e-4) $\to$ Constant(1.07e-4, 10.2T) $\to$ Cosine(3e-5, 7.5T) $\to$ Constant(3e-5, 5.5T)$_{\text{8K} \to \text{32K}}$ $\to$ Cosine(1e-5, 0.5T) \\
        \midrule
        \makecell[lt]{Pangu Ultra (2025/04) \\ \citep{yin2025panguultrapushinglimits}} & Warmup(1e-4) $\to$ Cosine(1e-5, 7.4T) $\to$ Constant(1e-5, 4.6T) $\to$ Cosine(7.5e-6, 0.8T)$_{\text{4K} \to \text{8K}}$ $\to$ Constant(7.5e-6, 0.4T)$_{\text{8K} \to \text{128K}}$  \\
        \midrule
        \makecell[lt]{MiniMax-Text-01 (2025/01) \\ \citep{minimax2025minimax01scalingfoundationmodels}} & Warmup(2e-4) $\to$ Constant(2e-4, 7.2T) $\to$ Constant(1.3e-4, 3.2T) $\to$ Exponential(3e-5, 1T)  \\
        \midrule
        \makecell[lt]{OLMo 2 32B (2024/12) \\ \citep{olmo20252olmo2furious}} & Warmup(6e-4) $\to$ Cosine(6e-5, 6.5T) $\to$ Linear(0) \\
        \midrule
        \makecell[lt]{DeepSeek-V3 (2024/12) \\ \citep{deepseekai2025deepseekv3technicalreport}} & Warmup(2.2e-4) $\to$ Constant(2.2e-4, 10T) $\to$ Cosine(2.2e-5, 4.3T) $\to$ Constant(2.2e-5, 333B)$_{\text{4K} \to \text{32K}}$ $\to$ Constant(7.3e-6, 167B)$_{\text{32K} \to \text{128K}}$ \\
        \midrule
        \makecell[lt]{Phi-4 14B (2024/12) \\ \citep{abdin2024phi4technicalreport}} & Warmup(3e-4) $\to$ Decay(${\eta}_{\min}$, 10T) \\
        \midrule
        \makecell[lt]{Yi-Lightning (2024/12) \\ \citep{wake2025yilightningtechnicalreport}} & Warmup(${\eta}_{\max}$) $\to$ Decay($\frac{1}{2}{\eta}_{\max}$) $\to$ Decay(${\eta}_{\min}$) \\
        \midrule
        \makecell[lt]{Hunyuan-Large (2024/11) \\ \citep{sun2024hunyuanlargeopensourcemoemodel}} & Warmup(${\eta}_{\max}$) $\to$ (Gradual)Decay(-) $\to$ Decay($\frac{1}{10}{\eta}_{\max}$, 5\%) \\
        \midrule
        \makecell[lt]{AFM 3B (2024/07) \\ \citep{gunter2024appleintelligencefoundationlanguage}} & Warmup(1e-2) $\to$ Cosine(5e-5, 6.3T); Warmup(3e-4) $\to$ Decay(3e-7, 1T) \\
        \midrule
        \makecell[lt]{Llama-3 405B (2024/07) \\ \citep{grattafiori2024llama3herdmodels}} & Warmup(8e-5) $\to$ Cosine(8e-7, 15.6T) $\to$ Decay(-, 800B)$_{\text{8K} \to \text{128K}}$ $\to$ Linear(0, 40M) \\
        \midrule
        \makecell[lt]{DeepSeek-V2 (2024/07) \\ \citep{grattafiori2024llama3herdmodels}} & Warmup(2.4e-4) $\to$ Constant(2.4e-4, 4.86T) $\to$ Constant(7.58e-5, 3.79T) $\to$ Decay(2.39e-5, 810B) \\
        \midrule
        \makecell[lt]{MiniCPM (2024/04) \\ \citep{hu2024minicpmunveilingpotentialsmall}} & Warmup(1e-2) $\to$ Constant(1e-2, 1T) $\to$ Exponential(${\eta}_{\min}$, 20B) \\
        \bottomrule
    \end{tabularx}
    \end{center}
\end{table}

Meanwhile, LR scheduling is the key to the overall performance optimization of the models. As illustrated in Table~\ref{tab:lrModels}, most mainstream models adopt a linear warm-up phase with a certain number of steps and a cosine decay phase. The WSD scheduler~\citep{hu2024minicpmunveilingpotentialsmall}, as a multi-stage LR scheduling, which maintains a constant LR in the initial stage and then linearly decays in the final stage, generates performance comparable to that of the cosine scheduling under an equivalent computational budget. Due to its excellent optimization characteristics, many subsequent models have adopted similar LR scheduling.
Notably, coupling the fast decay stage with high-quality data yields superior model convergence. Theoretical analysis suggests that the rapid convergence in the LR decay stage is due to the first-order directional derivative diminishing with each step, whereas the curvature of the loss function increases, close to the local optimum.
Subsequent research~\citep{hägele2024scalinglawscomputeoptimaltraining} has systematically validated WSD's equivalence to cosine schedules through scaling law analysis, and meanwhile revealed the rules of the LR schedule: the longer the annealing stage is within 20\% of the total token volume, the better. For shorter training cycles (210M/$<$5B tokens), 10-20\% decay phases outperform cosine schedules, while extended training (210M/20B tokens) achieves parity with merely 5\% decay. It also tests different decay functions and proves that the 1-sqrt function could achieve the best effect.

Since combining high-quality data with LR decay can better enhance the model performance, various LLMs have developed different multi-stage LR adjustment strategies based on the demands of the training stage. As illustrated in Table~\ref{tab:lrModels}, a prevailing trend is observed where models employ cosine or WSD during the pre-training stage to gradually reduce the LR, and subsequently, in the mid-training stage, there is a rapid decline in LR with high-quality data. With the popularity of this trend, some scholars are concerned that using a lower MAX LR in the mid-training stage may affect the convergence effect. In response, OctoThinker~\citep{wang2025octothinker} has attempted to use WSD in the mid-training stage and maintain a constant LR in the early stage of mid-training while adopting rapid decay with high-quality CoT data in the final stage, which achieves good results. Despite numerous attempts to refine these strategies in the mid-training stage, the complexity introduced by multiple variables, such as the starting and ending points of the LR schedule, Max LR, and End LR, necessitates further exploration.

\subsection{Core Cognitive Capabilities}

Core cognitive capabilities represent essential dimensions distilled from the general abilities of LLMs, serving as key areas of focus for advancing model performance during mid-training, including knowledge mastery and reasoning capabilities, covering areas such as mathematics and STEM.

\subsubsection{Knowledge Mastery}

Knowledge mastery positions LLMs as expansive knowledge repositories, with mid-training serving as a critical stage for further enhancing the model’s understanding of acquired knowledge and facilitating the acquisition of new information. Previous research has indicated that models exhibit insufficient mastery in domains such as mathematics and STEM~\citep{olmo20252olmo2furious, tencenthunyuanteam2025hunyuanturbosadvancinglargelanguage}, underscoring the importance of targeted improvements in areas where pre-trained models are weak due to natural data distribution biases.
Additionally, during the mid-training stage, probe experiments can be conducted to identify and address gaps in the model’s knowledge~\citep{zhang2025largescalediversesynthesismidtraining}. The primary focus is to deepen the model’s understanding of knowledge that may not be fully absorbed during pre-training while injecting new information. This can be achieved by mining and refining existing knowledge fragments within the pre-training corpus through quality filtering and synthetic data augmentation, integrating knowledge into QA formats~\citep{maini2024rephrasing}, and introducing new information by carefully collecting or generating domain-specific datasets that are absent from the original training data.

\begin{table}[t]
    \caption{Mid-training objectives of mainstream models. The checkmark indicates that the model explicitly mentions the corresponding mid-training optimization objective. Abbreviations: Reason. = Reasoning, Instruct. = Instruction Following, Long. = Long-context, Multi. = Multilingual.}
    \label{tab:targets}
    \begin{center}
    \setlength{\tabcolsep}{1mm}
    \small
    \begin{tabular}{l|cccccccc}
        \toprule
        \bf Model & \bf General & \bf Math & \bf Reason. & \bf Instruct. & \bf Code & \bf Agent & \bf Long. & \bf Multi.  \\
        \midrule
        \makecell[lt]{LongCat-Flash-Thinking (2025/09) \\ \citep{meituanlongcatteam2025longcatflashthinkingtechnicalreport}} & \CheckmarkBold & \CheckmarkBold & \CheckmarkBold & \CheckmarkBold & \CheckmarkBold & \CheckmarkBold & \CheckmarkBold \\
        % \midrule
        \makecell[lt]{LongCat-Flash (2025/09) \\ \citep{meituanlongcatteam2025longcatflashtechnicalreport}} & \CheckmarkBold & \CheckmarkBold &  & \CheckmarkBold & \CheckmarkBold & & \CheckmarkBold & \\
        \makecell[lt]{dots.llm1 (2025/05) \\ \citep{huo2025dotsllm1technicalreport}} & \CheckmarkBold & \CheckmarkBold & \CheckmarkBold & & \CheckmarkBold & & \CheckmarkBold & \CheckmarkBold \\
        % \midrule
        \makecell[lt]{MiMo (2025/05) \\ \citep{xiaomi2025mimounlockingreasoningpotential}} & \CheckmarkBold & \CheckmarkBold & \CheckmarkBold & & \CheckmarkBold & & \CheckmarkBold & \\
        \makecell[lt]{Qwen3 (2025/05) \\ \citep{yang2025qwen3technicalreport}} & \CheckmarkBold & \CheckmarkBold & \CheckmarkBold & \CheckmarkBold & \CheckmarkBold & & \CheckmarkBold & \CheckmarkBold \\
        \makecell[lt]{Pangu Ultra (2025/04) \\ \citep{yin2025panguultrapushinglimits}} & \CheckmarkBold & \CheckmarkBold & \CheckmarkBold & \CheckmarkBold & \CheckmarkBold & & \CheckmarkBold & \CheckmarkBold \\
        \makecell[lt]{Kimi K1.5 (2025/01) \\ \citep{kimiteam2025kimik15scalingreinforcement}} & \CheckmarkBold & \CheckmarkBold & \CheckmarkBold & \CheckmarkBold & \CheckmarkBold & & \CheckmarkBold & \CheckmarkBold \\
        % \midrule
        \makecell[lt]{OLMo 2 (2024/12) \\ \citep{olmo20252olmo2furious}} & \CheckmarkBold & \CheckmarkBold & & \CheckmarkBold & \CheckmarkBold & & & \\
        % \midrule
        \makecell[lt]{DeepSeek-V3 (2024/12) \\ \citep{deepseekai2025deepseekv3technicalreport}} & \CheckmarkBold & \CheckmarkBold & & & \CheckmarkBold & & \CheckmarkBold & \CheckmarkBold \\
        % \midrule
        \makecell[lt]{Phi-4 (2024/12) \\ \citep{abdin2024phi4technicalreport}} & \CheckmarkBold & \CheckmarkBold & \CheckmarkBold & \CheckmarkBold & \CheckmarkBold & \CheckmarkBold & \CheckmarkBold & \CheckmarkBold \\
        % \midrule
        \makecell[lt]{Yi-Lightning (2024/12) \\ \citep{wake2025yilightningtechnicalreport}} & \CheckmarkBold & \CheckmarkBold & \CheckmarkBold & \CheckmarkBold & \CheckmarkBold & & \CheckmarkBold & \CheckmarkBold \\
        \makecell[lt]{Hunyuan-Large (2024/11) \\ \citep{sun2024hunyuanlargeopensourcemoemodel}} & \CheckmarkBold & \CheckmarkBold & \CheckmarkBold & \CheckmarkBold & \CheckmarkBold & & \CheckmarkBold & \CheckmarkBold \\
        % \midrule
        \makecell[lt]{AFM (2024/07) \\ \citep{gunter2024appleintelligencefoundationlanguage}} & \CheckmarkBold & \CheckmarkBold & & & \CheckmarkBold & & \CheckmarkBold & \\
        % \midrule
        \makecell[lt]{Llama-3 (2024/07) \\ \citep{grattafiori2024llama3herdmodels}} & \CheckmarkBold & \CheckmarkBold & & & \CheckmarkBold & & \CheckmarkBold & \CheckmarkBold \\
        % \midrule
        \makecell[lt]{Nemotron-4 (2024/06) \\ \citep{nvidia2024nemotron4340btechnicalreport}} & \CheckmarkBold & & & & \CheckmarkBold & & & \CheckmarkBold \\
        \makecell[lt]{MiniCPM (2024/04) \\ \citep{hu2024minicpmunveilingpotentialsmall}} & \CheckmarkBold & \CheckmarkBold & & \CheckmarkBold & \CheckmarkBold & & & \\
        \bottomrule
    \end{tabular}
    \end{center}
\end{table}

\subsubsection{Reasoning Capabilities}

Systematic enhancement of reasoning capabilities is one of the central focuses during mid-training, as shown in Table \ref{tab:targets}. It not only targets the direct improvement of core cognitive dimensions—such as logical deduction, causal inference, and multi-step problem solving, but also lays a robust foundation for subsequent RL development~\citep{wang2025octothinker}. 

To achieve this dual objective, prevailing methodologies strategically enrich reasoning-intensive datasets through large-scale synthesis of high-quality reasoning samples, including CoT sequences, mathematical proof chains, and structured decision trajectories. For instance, CoTP~\citep{zhang2025expandingreasoningpotentialfoundation} proposes leveraging data abundant in diverse, high-value reasoning patterns to expand the model’s reasoning potential, selected by a dual-granularity DTW matching algorithm based on reasoning pattern chains and token entropy chains. 
Some methodologies further dynamically adjust the data mixing ratios, progressively increasing the proportion of reasoning data without compromising general functionality~\citep{wake2025yilightningtechnicalreport, olmo20252olmo2furious}. 
Such data interventions synergize with innovative training strategies, including long-context extension mechanisms that scaffold complex reasoning chains~\citep{hu2024minicpmunveilingpotentialsmall}. As synthetic data can embed diverse, learnable reasoning patterns during mid-training~\citep{abdin2024phi4technicalreport}, this intrinsic property positions reasoning optimization as the critical nexus bridging foundation models with downstream capabilities.

\subsection{Task Execution Capabilities}

Task execution capabilities highlight the pivotal role of mid-training as a warm-up stage for subsequent post-training. During mid-training, the model's potential to follow instructions, perform coding tasks, and exhibit agentic behaviors can be systematically enhanced.

\subsubsection{Instruction Following}

Instruction following is designed to enhance the model’s capacity to accurately adhere to given instructions, which is a critical skill for the development of post-training capabilities. Prior research~\citep{wang2025octothinker} has demonstrated that even a small proportion of instruction-following data during mid-training can significantly improve the model’s ability to handle complex instructions. As shown in the table, an increasing number of mainstream models have begun to incorporate a certain amount of high-quality instruction data during the mid-training stage.

\subsubsection{Coding}

The objective of the coding domain is to boost the model's proficiency in code generation, completion, fixing, and reasoning. As shown in Table~\ref{tab:targets}, recent advancements have focused on synthesizing large volumes of high-quality code data and employing upsampling techniques to further improve coding capabilities during the mid-training stage. This approach ensures that models are exposed to diverse and complex programming scenarios, thereby strengthening their overall performance in code-related tasks and laying a solid foundation for subsequent development.

\subsubsection{Agent}

Agent capability is a critical area of optimization for downstream applications of large language models, with tool usage currently standing out as the most prominent and actively researched aspect within this domain. The ability of foundational models to effectively utilize external tools is essential for enabling LLMs to tackle complex problems in real-world environments, where tasks often require interaction with diverse systems and dynamic resources. However, natural data related to tool usage remains inherently scarce, and the diverse, complex nature of real-world environments introduces substantial challenges, such as high trajectory sampling costs and difficulties in verifying output correctness. To address these challenges, Kimi K2~\citep{kimiteam2025kimik2openagentic} employs a task synthesis pipeline, achieving outstanding performance across multiple tool-use evaluation benchmarks. This highlights the potential of synthetic data in advancing tool usage capabilities. Building large-scale datasets specifically focused on tool usage and increasing the proportion of such data during the mid-training stage is thus a vital strategy for further improving the agent capabilities of foundational models and empowering LLMs to solve increasingly complex problems in real environments.

\subsection{Extension Capabilities}

\subsubsection{Long Context Extension}

% [TODO: table]
\begin{table}[t]
\centering
\caption{Comparison of long context extension strategies across 13 mainstream LLMs. The Tokens column indicates training tokens for the extension phase only. N/S denotes ``Not Specified.''}
\label{tab:context_extension}
\begin{center}
\setlength{\tabcolsep}{1mm}
\small
\begin{tabular}{l|cccccc}

\toprule
\textbf{Model} & \textbf{Max Length} & \textbf{Stages} & \textbf{Tokens} & \textbf{RoPE Base} & \textbf{Method} & \textbf{Long Data} \\
\midrule
\makecell[lt]{LongCat-Flash (2025/09) \\ \citep{meituanlongcatteam2025longcatflashtechnicalreport}} & \makecell[ct]{8K$\rightarrow$32K\\$\rightarrow$128K} & 2 & 100B & \makecell[ct]{1M$\rightarrow$5M\\$\rightarrow$10M} & ABF & 25\% \\
\midrule
\makecell[lt]{GLM-4.5 (2025/08) \\ \citep{5team2025glm45agenticreasoningcoding}} & \makecell[ct]{4K$\rightarrow$32K\\$\rightarrow$128K} & 2 & 100B & \makecell[ct]{10K$\rightarrow$1M} & ABF & N/S \\
\midrule
\makecell[lt]{ERNIE 4.5 (2025/06) \\ \citep{ernie2025technicalreport}} & \makecell[ct]{8K$\rightarrow$32K\\$\rightarrow$128K} & 2 & N/S & \makecell[ct]{100K$\rightarrow$160K\\$\rightarrow$500K} & ABF & N/S \\
\midrule
\makecell[lt]{dots.llm1 (2025/05) \\ \citep{huo2025dotsllm1technicalreport}} & $\rightarrow$32K & 1 & 128B & N/S & ABF & N/S \\
\midrule
\makecell[lt]{Qwen3 (2025/05) \\ \citep{yang2025qwen3technicalreport}} & $\rightarrow$32K (128K) & 1 & 100B & 10K$\rightarrow$1M & \makecell[ct]{ABF+YARN\\+DCA} & \makecell[ct]{75\%\\(16--32K)} \\
\midrule
\makecell[lt]{Pangu-Ultra (2025/04) \\ \citep{yin2025panguultrapushinglimits}} & \makecell[ct]{8K$\rightarrow$32K\\$\rightarrow$128K} & 2 & N/S & 1.6M$\rightarrow$25.6M & ABF & N/S \\
\midrule
\makecell[lt]{DeepSeek-V3 (2024/12) \\ \citep{deepseekai2025deepseekv3technicalreport}} & \makecell[ct]{4K$\rightarrow$32K\\$\rightarrow$128K} & 2 & \makecell[ct]{N/S\\(2K steps)} & N/S & YaRN & N/S \\
\midrule
\makecell[lt]{Phi-4 (2024/12) \\ \citep{abdin2024phi4technicalreport}} & 4K$\rightarrow$16K & 1 & 250B & 250K & ABF & 30\% \\
\midrule
\makecell[lt]{Yi-Lightning (2024/12) \\ \citep{wake2025yilightningtechnicalreport}} & $\rightarrow$64K & Multi & 20B & N/S & ABF & N/S \\
\midrule
\makecell[lt]{Hunyuan-Large (2024/11) \\ \citep{tencenthunyuanteam2025hunyuanturbosadvancinglargelanguage}} & $\rightarrow$32K$\rightarrow$256K & 2 & 20B & 1B & ABF & 25\% \\
\midrule
\makecell[lt]{AFM (2024/07) \\ \citep{gunter2024appleintelligencefoundationlanguage}} & $\rightarrow$32K & 1 & 100B & 500K$\rightarrow$6.3M & ABF & N/S \\
\midrule
\makecell[lt]{Llama-3 (2024/07) \\ \citep{grattafiori2024llama3herdmodels}} & 8K$\rightarrow$128K & 6 & ${\sim}800$B & 500K & ABF & N/S \\
\midrule
\makecell[lt]{MiniCPM (2024/04) \\ \citep{hu2024minicpmunveilingpotentialsmall}} & 4K$\rightarrow$128K & 2 & N/S & N/S & \makecell[ct]{ABF+\\NTK-RoPE} & 44\% \\
\bottomrule
\end{tabular}
\end{center}
\end{table}

The long context extension is designed to increase the context length that a model can effectively handle, thereby improving its ability to comprehend lengthy documents. In this section, we examine the long context extension strategies employed by mainstream LLMs, analyzing their technical approaches across position encoding modifications, training methodologies, and data composition strategies. We provide a comprehensive comparison in Table~\ref{tab:context_extension} that reveals common practices in extending context windows from standard lengths (4K-8K tokens) to ultra-long contexts (up to 256K tokens).

\textbf{Position Encoding Modifications.}
All surveyed models utilize Rotary Position Embedding (RoPE)~\citep{roformer} as their foundation but employ different scaling strategies. The most common approach involves Adaptive Base Frequency (ABF)~\citep{xiong-etal-2024-abf}, which can be categorized into three strategies: \textit{conservative scaling} (Phi-4: 250K for 16K context), \textit{moderate scaling} (AFM: 6.3M for 32K; Qwen3: 1M for 32K), and \textit{aggressive scaling} (Hunyuan: 1B for 256K; Pangu-Ultra: 25.6M for 128K). 
Advanced techniques include NTK-Aware RoPE Scaling~\citep{ntkaware2023} utilized by MiniCPM for extreme extensions (32K$\rightarrow$128K) and YaRN~\citep{peng2024yarn} adopted by DeepSeek-V3 and Qwen3. Notably, Qwen3 combines multiple techniques (ABF, YaRN, DCA~\citep{10.5555/3692070.3692130dca}) to achieve $4\times$ inference extension beyond training length.

\textbf{Training Strategies.}
Progressive, multi-stage extension represents the dominant paradigm. Two-stage approaches are most prevalent (DeepSeek-V3, Hunyuan, Pangu-Ultra, ERNIE 4.5, GLM-4.5, LongCat-Flash: base$\rightarrow$32K$\rightarrow$128K), while Llama-3 implements the most granular six-stage extension (8K$\rightarrow$128K).
Token budgets vary significantly across models: \textit{ultra-large scale} (Llama-3: ${\sim}800$B, Phi-4: 250B), \textit{large scale} (dots.llm1: 128B, AFM: 100B), \textit{medium scale} (GLM-4.5: 100B, LongCat-Flash: 100B), and \textit{small scale} (Hunyuan: 20B, Yi-lightning: 20B).
Models consistently reduce the batch size proportionally with context length increases. For instance, DeepSeek-V3 and Pangu-Ultra both implement $4\times$ reductions (1920$\rightarrow$480 and 384$\rightarrow$96, respectively) when extending from 32K to 128K tokens.

\textbf{Data Composition.}
% Models converge on 25\% to 44\% long-context data mixed with standard-length data: MiniCPM (44\% long), Phi-4 (30\% long), and Hunyuan and LongCat-Flash (25\% long). Data sourcing philosophies diverge significantly---Hunyuan relies solely on naturally occurring long-context data (books and code repositories), while MiniCPM and AFM augment with synthetic long-form question-answering data. Phi-4 demonstrates that inherently long documents outperform artificially concatenated sequences.
Data composition critically balances capability extension with stability preservation. Models converge on maintaining 70-75\% short-context data mixed with 25-30\% long-context sequences, establishing an empirically validated sweet spot. The superiority of naturally long documents over artificially concatenated sequences has been empirically confirmed, with Phi-4's direct ablations demonstrating substantially better performance from inherently long sources. Consistently emphasized natural sources include books and novels, repository-level code, and academic papers. Despite natural data advantages, synthetic long QA data plays a strategic role in enhancing context-aware task performance (MiniCPM, AFM, Yi-Lightning), serving as augmentation rather than replacement.
Quality curation emerges as critical: Phi-4 filters for samples exceeding 8K tokens and upweights data ${\geq}16$K tokens, LongCat-Flash applies multi-stage filtering to repository-level code, GLM-4.5 and Llama-3 upsample high-quality sources. Dots.llm1 employs the Untie the Knots (UtK) strategy~\citep{tian-etal-2025-untie-utk}, which chunks documents and trains reconstruction from shuffled segments rather than requiring extensive long-document datasets.

Key observations include: (1)~two-stage extension represents the dominant pattern, (2)~RoPE base frequency scaling correlates with target context length, (3)~the 25--30\% long-context data range suggests an empirically validated sweet spot, and (4)~token budgets span two orders of magnitude (20B--800B).
These findings establish that effective context extension requires careful orchestration of position encoding modifications, progressive training schedules, balanced data mixing (25--30\% long-context data), and length-stratified upsampling and rigorous quality filtering ensuring both long-context capability and short-context performance preservation. The diversity in token budgets (20B--800B) and approaches (natural versus synthetic data, single versus multi-stage training) suggests multiple viable paths depending on model scale, target context length, and computational constraints.

\subsubsection{Language Extension}

Language extension aims to improve model performance in multilingual environments, with a particular focus on low-resource languages. Most models address multilingual objectives by proactively enlarging their vocabulary size, as evidenced by the Yi-Lightning model~\citep{wake2025yilightningtechnicalreport}, which expanded its vocabulary to 100,352 tokens to enhance multilingual support. In addition to vocabulary expansion, the process involves collecting or synthesizing diverse multilingual datasets and gradually increasing their sampling ratio throughout mid-training. This strategy enables the model to effectively learn and process multiple languages, including those with scarce resources, thereby significantly extending its applicability and utility in multilingual scenarios.

\section{Conclusion}

In conclusion, mid-training has emerged as a pivotal and distinct stage, bridging pre-training and post-training.  By systematically integrating objective-driven interventions, such as data curation, training strategies, and model architecture optimization, mid-training enables the efficient enhancement of specific capabilities like mathematics, coding, reasoning, and long-context extension, while simultaneously preserving the general abilities acquired during pre-training. Empirical evidence consistently demonstrates that mid-training contributes to steeper performance improvements and greater training efficiency, owing to its intermediate data and computational requirements and its strategic positioning within the overall training pipeline. This survey provides a comprehensive taxonomy and operational insight into mid-training, formally delineating its conceptual boundaries and optimization frameworks, laying the foundation for more systematic research and development in the objective-oriented advancement of LLMs.

\bibliography{main}
\bibliographystyle{main}

% \appendix
% \section{Appendix}
% You may include other additional sections here.

\end{document}